\documentclass[conference]{IEEEtran}
\IEEEoverridecommandlockouts
\usepackage{cite}
\usepackage{url}
\makeatletter
\newcommand*{\rom}[1]{\expandafter\@slowromancap\romannumeral #1@}
\makeatother
\usepackage{amsmath,amssymb,amsfonts}
\usepackage{algorithmic}
\usepackage{graphicx}
\usepackage{textcomp}
\usepackage{xcolor}
\def\BibTeX{{\rm B\kern-.05em{\sc i\kern-.025em b}\kern-.08em
    T\kern-.1667em\lower.7ex\hbox{E}\kern-.125emX}}
\begin{document}

\title{Continuous Speech Recognition using EEG and Video\\
{
}
\thanks{*Equal contribution.}
}

\author{\IEEEauthorblockN{Gautam Krishna}
\IEEEauthorblockA{\textit{Brain Machine Interface Lab} \\
\textit{The University of Texas at Austin}\\
Austin, Texas \\
}
\and
\IEEEauthorblockN{Mason Carnahan*}
\IEEEauthorblockA{\textit{Brain Machine Interface Lab} \\
\textit{The University of Texas at Austin}\\
Austin, Texas \\
}
\and
\IEEEauthorblockN{Co Tran*}
\IEEEauthorblockA{\textit{Brain Machine Interface Lab} \\
\textit{The University of Texas at Austin}\\
Austin, Texas \\
}
\and
\IEEEauthorblockN{Ahmed H Tewfik}
\IEEEauthorblockA{\textit{Brain Machine Interface Lab} \\
\textit{The University of Texas at Austin}\\
Austin, Texas  \\
}
}

\maketitle

\begin{abstract}
In this paper we investigate whether electroencephalography (EEG) features can be used to improve the performance of continuous visual speech recognition systems. We implemented a connectionist temporal classification (CTC) based end-to-end automatic speech recognition (ASR) model for performing recognition. Our results demonstrate that EEG features are helpful in enhancing the performance of continuous visual speech recognition systems. 
\end{abstract}

\begin{IEEEkeywords}
electroencephalograpgy (EEG), speech recognition, deep learning, CTC, technology accessibility, computer vision  
\end{IEEEkeywords}

\section{Introduction}
In recent years there has been lot of interesting work done in the fields of lip reading and audio visual speech recognition. In \cite{assael2016lipnet} authors demonstrated end-to-end sentence level lip reading and in \cite{afouras2018deep} authors demonstrated deep learning based end-to-end audio visual speech recognition. Similarly there has been lot of new results published in the field of speech recognition using bio signals, mainly using electrocorticography (ECoG) and electroencephalography (EEG).  ECoG is an invasive way of measuring electrical activity of human brain where a subject need to undergo a brain surgery to get ECoG electrodes implanted. On the other hand EEG is a non invasive way of measuring electrical activity of human brain where signals are recorded by placing EEG sensors on the scalp of the subject. 
In \cite{krishna2019speech} authors demonstrated isolated speech recognition using EEG and combination of EEG, acoustic features on a limited English vocabulary of four words and five vowels. In \cite{krishna20,krishna2019state} authors demonstrated continuous speech recognition using EEG features in clean and noisy environments. In \cite{krishna2019speech} authors demonstrated that EEG features are helpful in improving the robustness of automatic speech recognition (ASR) systems operating in noisy environments. In this paper we investigate whether EEG features are more helpful than acoustic features to improve the performance of continuous visual speech recognition systems.  

Given the limited amount of data we had in our hand we implemented our own custom model for performing recognition instead of using state-of-the art computer vision feature extraction network architectures like Resnet \cite{he2016deep} or VGG net \cite{simonyan2014very}. We also avoided trying transfer learning or fine-tuning pre-trained Resnet or VGG mainly because of our computing hardware memory limitations and the main goal of the work explained in this paper was to investigate the feasibility of using EEG features to enhance the performance of continuous visual speech recognition systems. The goal of this work was not to outperform the performance of current state-of-art visual speech recognition systems. 

Our results demonstrate that EEG features are in fact helpful in improving the performance of continuous visual speech recognition systems. We demonstrate our results for a limited English vocabulary consisting of 30 unique sentences.

\section{Visual Speech Recognition System Models}
Figure 1 explains the architecture  of the recognition model used for recognition of combination of video and EEG data, video and acoustic data and combination of video,acoustic and EEG data. The part of the network used for extracting features from EEG and acoustic data consists of three layers of gated recurrent unit (GRU) \cite{chung2014empirical} with 128, 64 and 32 hidden units respectively. Each GRU layer included a dropout regularization \cite{srivastava2014dropout} with dropout rate 0.1. This part of the network can take EEG or acoustic features or concatenation of EEG and acoustic features as input depending on how the model is trained. 
The part of the network used for extracting features from video frames consisted of two dimensional convolutional network layers and two dimensional max pooling layer. We used two convolutional network layers and one max pooling layer.  The convolutional layers had 100 filters with ReLU \cite{agarap2018deep} activation function and a kernel size of (1,3) and the max pooling layer had a pool size of (1,2). The output of the max pool layer is flattened and reshaped in order to concatenate it with the features extracted by the other part of the network described before. The concatenated features are fed into a temporal convolutional network (TCN) \cite{bai2018empirical} layer consisting of 32 filters, whose output is fed into the decoder of the connectionist temporal classification (CTC) \cite{graves2014towards,krishna2019state,krishna20} network. The decoder of the CTC network consists of combination of a dense layer which performs affine transformation and a softmax activation. The output of the encoder is fed into the decoder of the CTC network at every time step. The number of time steps of the encoder is same as the product of sampling frequency of the input features and sequence length. There was no fixed time step value since different subjects spoke with different rate of speech. We used a character based CTC model in this work. The model was predicting a character at every time step. 

The details of the CTC loss function are covered in \cite{graves2014towards,krishna2019state,krishna20}. During inference time we used a combination of CTC beam search decoder and an external 4-gram language model, known as shallow fusion \cite{toshniwal2018comparison}. Figure 2 shows the architecture of the model used for performing speech recognition using only video features. It is very similar to the model explained in Figure 1 except it doesn't contain additional network layers to extract acoustic or EEG features. 

Both the models were trained for 120 epochs using adam \cite{kingma2014adam} optimizer and the batch size was set to 100. Validation split was 0.1.  Figure 3 shows the CTC loss convergence for the model during training. All the scripts were written using keras and tensorflow 2.0 deep learning framework.

\begin{figure}[h]
\begin{center}
\includegraphics[height=8.5cm, width=\linewidth,trim={0.1cm 0.1cm 0.1cm 0.1cm}]{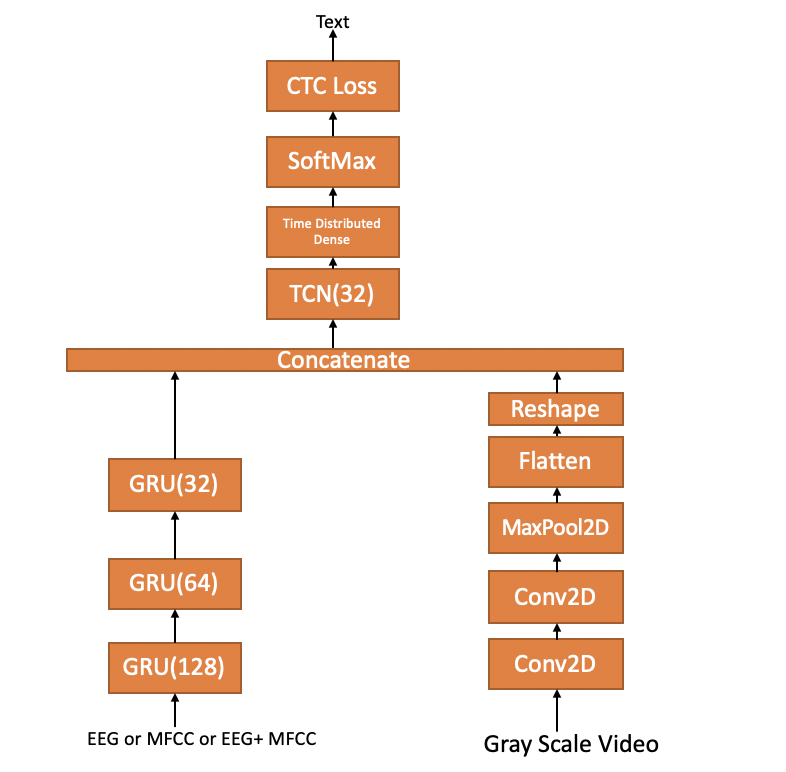}
\caption{Video-EEG Fusion Recognition Model} 
\label{1vsall}
\end{center}
\end{figure}

\begin{figure}[h]
\begin{center}
\includegraphics[height=8.5cm, width=\linewidth,trim={0.1cm 0.1cm 0.1cm 0.1cm}]{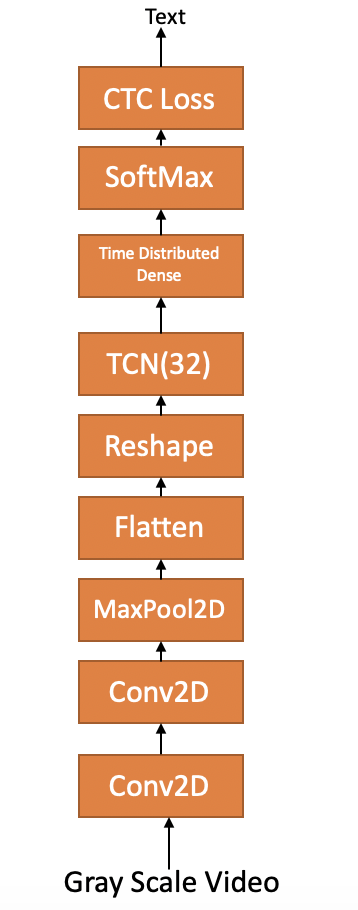}
\caption{Video Recognition Model} 
\label{1vsall}
\end{center}
\end{figure}

\begin{figure}[h]
\includegraphics[height=4.5cm, width=0.4
\textwidth,trim={0.1cm 0.1cm 0.1cm 0.1cm},clip]{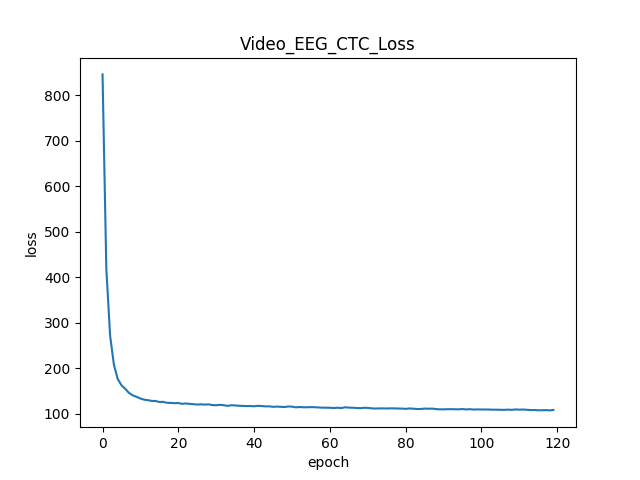}
\caption{CTC loss convergence for Video-EEG fusion model}
\label{1vsall}
\end{figure}

\begin{figure}[h]
\begin{center}
\includegraphics[height=3cm,width=0.25\textwidth,trim={1cm 1cm 1cm 0.1cm},clip]{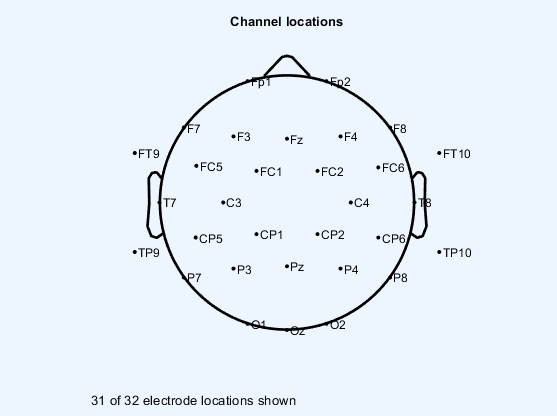}
\caption{EEG channel locations for the cap used in our experiments} 
\label{1vsall}
\end{center}
\end{figure}

\section{Design of Experiments for building the database}
Seven male UT Austin graduate students in their early to mid twenties took part in the speech-EEG-Video experiment. 
Each subject was asked to speak the first 30 English sentences from USC-TIMIT data base \cite{narayanan2014real} and their simultaneous speech, EEG and video was recorded.
The sentences were shown to subjects on a computer screen and they read out loud the sentences. The data was recorded in presence of a background noise of 65dB. The music played from our lab computer was used as the source of noise. Then each subject was asked to repeat the same experiment two more times, thus we had 90 speech-eeg-video recordings from each subject. 

We used Brain product's ActiChamp EEG amplifier. Our EEG cap had 32 wet EEG electrodes including one electrode as ground as shown in Figure 4. We used EEGLab \cite{delorme2004eeglab} to obtain the EEG sensor location mapping. It is based on standard 10-20 EEG sensor placement method for 32 electrodes.

Data from first 6 subjects was used as the training set and the last subject data was used as the test set. 

\begin{table*}[!ht]
\centering
\begin{tabular}{|l|l|l|l|l|l|l|l|l|l|}
\hline
\textbf{\begin{tabular}[c]{@{}l@{}}Total \\ Number\\ of \\ Sentences\end{tabular}} & \textbf{\begin{tabular}[c]{@{}l@{}}Number\\  of\\ Unique\\ Sentences\\ Contained\end{tabular}} & \multicolumn{1}{c|}{\textbf{\begin{tabular}[c]{@{}c@{}}Total \\ Number\\ of \\ words\\  Contained\end{tabular}}} & \multicolumn{1}{c|}{\textbf{\begin{tabular}[c]{@{}c@{}}Number\\ of\\ Unique\\ words\\ Contained\end{tabular}}} & \textbf{\begin{tabular}[c]{@{}l@{}}Number\\ of Letters\\ Contained\end{tabular}} & \textbf{\begin{tabular}[c]{@{}l@{}}Video\\ WER\\ (\%)\end{tabular}} & \multicolumn{1}{c|}{\textbf{\begin{tabular}[c]{@{}c@{}}Video\\ +\\ MFCC\\ WER\\ (\%)\end{tabular}}} & \multicolumn{1}{c|}{\textbf{\begin{tabular}[c]{@{}c@{}}Video\\ +\\ EEG\\ WER\\ (\%)\end{tabular}}} & \multicolumn{1}{c|}{\textbf{\begin{tabular}[c]{@{}c@{}}Video\\ +\\ EEG\\ +\\ MFCC\\ WER\\ (\%)\end{tabular}}} & \textbf{\begin{tabular}[c]{@{}l@{}}MFCC\\ WER\\ (\%)\end{tabular}} \\ \hline
90                                                                                 & 30                                                                                             & 552                                                                                                              & 153                                                                                                            & 2598                                                                             & 96.93                                                               & 85.23                                                                                               & 84.60                                                                                              & \textbf{83.74}                                                                                                & 86.45                                                              \\ \hline
\end{tabular}
\caption{WER on Test Set}
\end{table*}

\section{Feature extraction and preprocessing details}
We followed the same EEG and speech preprocessing methods used by authors in \cite{krishna2019speech,krishna20}. 
EEG signals were sampled at 1000Hz and a fourth order IIR band pass filter with cut off frequencies 0.1Hz and 70Hz was applied. A notch filter with cut off frequency 60 Hz was used to remove the power line noise.
EEGlab's \cite{delorme2004eeglab} Independent component analysis (ICA) toolbox was used to remove other biological signal artifacts like electrocardiography (ECG), electromyography (EMG), electrooculography (EOG) etc from the EEG signals. 
We extracted five statistical features for EEG, namely root mean square, zero crossing rate,moving window average,kurtosis and power spectral entropy \cite{krishna2019speech,krishna20}. So in total we extracted 31(channels) X 5 or 155 features for EEG signals.The EEG features were extracted at a sampling frequency of 100Hz for each EEG channel.

The recorded speech signal was sampled at 16KHz frequency. We extracted Mel-frequency cepstrum coefficients (MFCC) as features for speech signal.
We extracted MFCC features of dimension 13. The MFCC features were also sampled at 100Hz, same as the sampling frequency of EEG features. 

We extracted 100 frames per second from the recorded video. We used YOLO\cite{redmon2016you} object recognition model to perform face recognition from the extracted video frames. Figure 6 shows a raw extracted RGB video frame and Figure 7 shows the corresponding face frame extracted using YOLO. The maximum x dimension value in our extracted face data set was 426 and maximum y dimension value in our extracted face data set was 381. Z was of dimension 3 (RGB).  Our initial plan was to perform experiments using RBG frames but we were constrained by memory requirements of our computing hardware, hence we transformed the RGB face frames to gray scale and resized all the gray scale face frames to a dimension of 100 X 100.  Figure 8 shows the corresponding gray scale resized face frame. We used python imaging library (PIL) for resizing the images. We tried extracting lip or mouth frames from the gray scale face frames using DLib and iBug face landmark predictor with 68 landmarks \cite{sagonas2013300} but the iBug face landmark predictor was not able to detect mouth or lips for all our face frames in the data set, possibly because of the EEG cap worn by the subjects causing the iBug face landmark predictor to give in-accurate mouth predictions. Figure 9 shows some of the lip frames extracted using iBug face landmark predictor. Because of the missing lip frames, we fed only the gray scale face frames to the model during training and test time. We recommend researchers to use RGB frames and use three dimensional convolutional and max pooling layers instead of two dimensional layers in the models described in Figures 1 and 2, if sufficient computing resources are available. 
\begin{figure}[h]
\centering
\includegraphics[height=5cm, width=0.4
\textwidth,trim={0.1cm 0.1cm 0.1cm 0.1cm},clip]{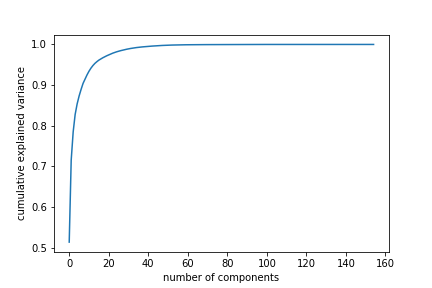}
\caption{Explained variance plot}
\label{1vsall}
\end{figure}

\begin{figure}[h]
\centering
\includegraphics[height=5cm, width=0.4
\textwidth,trim={0.1cm 0.1cm 0.1cm 0.1cm},clip]{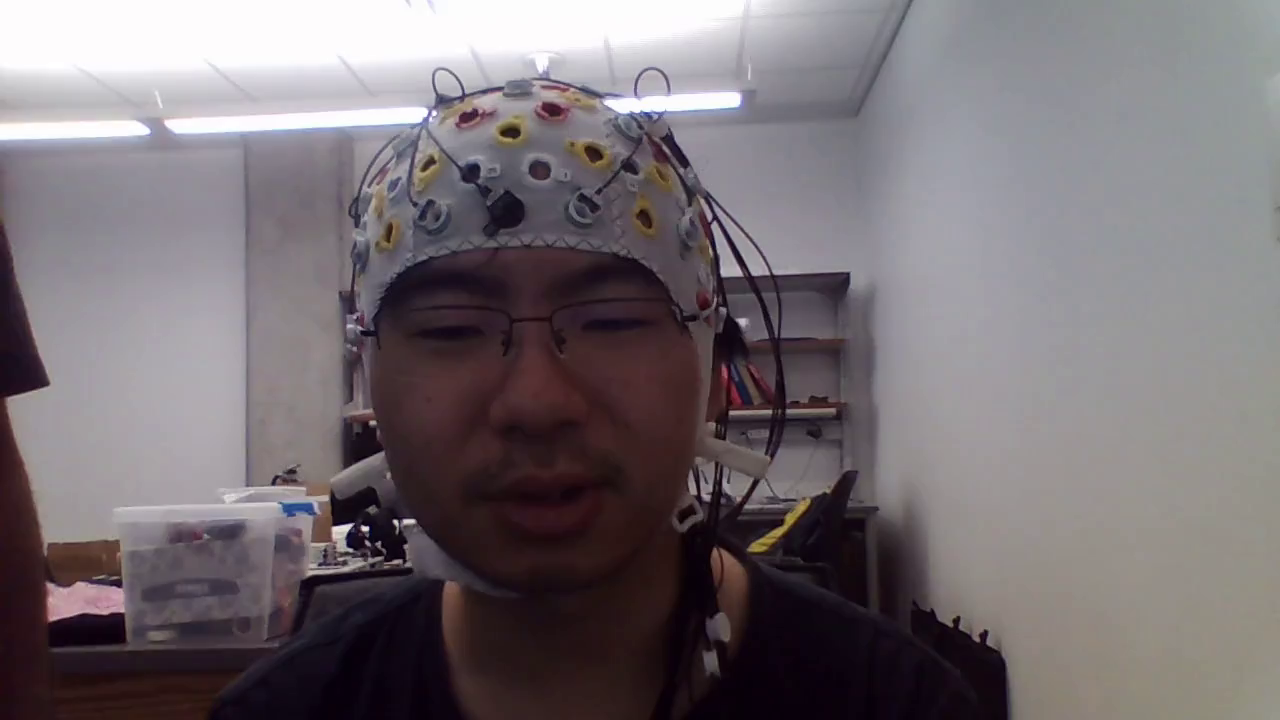}
\caption{Extracted raw video frame}
\label{1vsall}
\end{figure}

\begin{figure}[h]
\centering
\includegraphics[height=5cm, width=0.4
\textwidth,trim={0.1cm 0.1cm 0.1cm 0.1cm},clip]{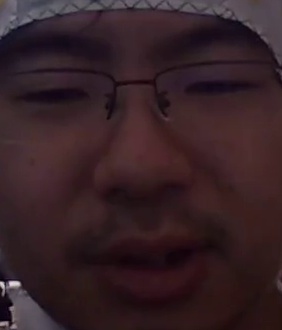}
\caption{Extracted RGB face frame from the raw video frame}
\label{1vsall}
\end{figure}

\begin{figure}[h]
\centering
\includegraphics[height=5cm, width=0.4
\textwidth,trim={0.1cm 0.1cm 0.1cm 0.1cm},clip]{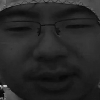}
\caption{RGB face frame resized and converted to gray scale}
\label{1vsall}
\end{figure}

\begin{figure}[h]
\centering
\includegraphics[height=6cm, width=0.55
\textwidth,trim={0.1cm 0.1cm 0.1cm 0.1cm},clip]{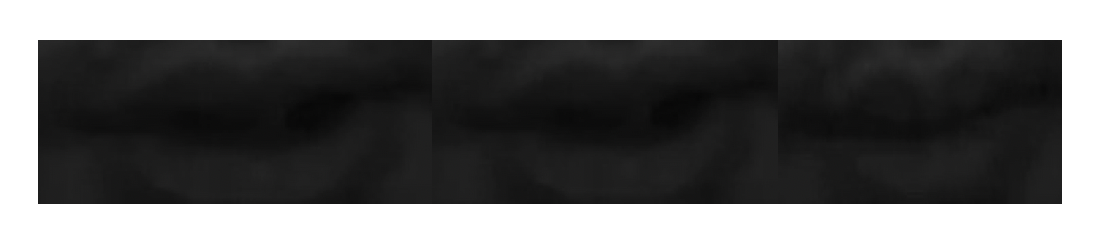}
\caption{gray scale lip frames extracted from gray scale face frames}
\label{1vsall}
\end{figure}

\section{EEG Feature Dimension Reduction Algorithm Details}
After extracting EEG and acoustic features as explained in the previous section, we used Kernel Principle Component Analysis (KPCA) \cite{mika1999kernel} to denoise the EEG feature space as explained by authors in \cite{krishna20,krishna2019speech}. 
We reduced the 155 EEG features to a dimension of 30 by applying KPCA for both the data sets. We plotted cumulative explained variance versus number of components to identify the right feature dimension as shown in Figure 5. We used KPCA with polynomial kernel of degree 3 \cite{krishna2019speech,krishna20}. 
\section{Results}

We used word error rate (WER) as performance metric of the model during test time. Table 1 shows the results obtained during test time. Table 1 shows the average WER on test set for various feature set inputs. The results demonstrate that EEG features are more helpful than acoustic features to improve the performance of continuous visual speech recognition systems operating in noisy environments. Using all modalities ( acoustic, EEG,  Video) gave the highest test time performance or the lowest word error rate on test set. For obtaining results shown in Table 1, faces frames were fed into the model.  We also tried performing experiments by using combination of lip frames and face frames (where we kept face frames when the iBug face landmark predictor failed to detect lip or mouth) but the test time results were worse with model giving a higher WER of 97.01 \% only with video data, hence we didn't perform more experiments with combination of face and lip frames. 
\section{Conclusion and Future work}
In this paper we demonstrated the feasibility of using EEG features to improve visual speech recognition systems operating in noisy environments. We validated our results on a test set vocabulary consisting of a total of 90 English sentences or 30 unique English sentences. 
For future work we would like to build a much larger data set and validate our results on a larger corpus. We would also to carry out experiments using RBG frames, do fine tuning on state-of-the art computer vision networks to extract better features from face frames, incorporate optical flow features for our future work in order to improve our current results.

We encourage other researchers to put a joint effort in building a state of the art Speech-Video-EEG data base to help advance research in this area.

\section{Acknowledgement} 
We would like to thank Kerry Loader and Rezwanul Kabir from Dell, Austin, TX for donating us the GPU to train the models used in this work.

The first author would like to thank Prof Yann Soullard from Université De Rennes 2 for his crucial help with debugging the CTC model code.

\bibliographystyle{IEEEtran}

\bibliography{refs}
\end{document}